\RequirePackage[hyphens]{url}
\documentclass{jdmdh}
\usepackage{array}
\usepackage{pgfplots}
\usepackage{cjhebrew}

\pgfplotsset{compat=1.18}
\titlespacing*{\section}
{0pt}{2.5ex plus 1ex minus .2ex}{1.3ex plus .2ex}
\titlespacing*{\subsection}
{0pt}{1.0ex plus 1ex minus .2ex}{1.3ex plus .2ex}
\titlespacing*{\subsubsection}
{0pt}{1.0ex plus 1ex minus .2ex}{1.3ex plus .2ex}

\title{Style Classification of Rabbinic Literature for\\ Detection of Lost Midrash Tan\d{h}uma Material}
\author[1]{Shlomo Tannor}
\author[1]{Nachum Dershowitz}
\author[2]{Moshe Lavee}
\affil[1]{Tel Aviv University, Israel}
\affil[2]{Haifa University, Israel}

\corrauthor{Shlomo Tannor}{shlomotannor@mail.tau.ac.il}

\begin{document}

\maketitle

\abstract{Midrash collections are complex rabbinic works that consist of text in multiple languages, which evolved through long processes of unstable oral and written transmission.
Determining the origin of a given passage in such a compilation is not always straightforward and is often a matter of dispute among scholars, yet it is essential for scholars' understanding of the passage and its relationship to other texts in the rabbinic corpus. 

To help solve this problem, we propose a system for classification of rabbinic literature based on its style, leveraging recent advances in natural language processing for Hebrew texts. Additionally, we demonstrate how this method can be applied to uncover lost material from a specific midrash genre, Tan\d{h}uma-Yelammedenu, that has been preserved in later anthologies.}


\section{Introduction}

Midrash, an integral genre within Jewish literature, encompasses a range of interpretative and narrative texts that seek to explore and expound upon the meanings of biblical scriptures. These texts incorporate a rich mix of legal, ethical, and philosophical discussions, allegories, parables, and homilies, offering deeper insights into the religious passages they explore.

Midrash anthologies are compilations of these interpretive works. They're inherently complex, containing text in multiple languages, written by different authors over several generations and geographic locations. The anthologists would often merge, quote, or paraphrase earlier sources. This process creates a composite that can make it difficult for scholars to discern the individual components. The identification of sections from particular sources within these anthologies can illuminate various scholarly debates, thereby enriching our understanding of the historical evolution of the rabbinic corpus.

The prospect of automated analysis and classification of rabbinic texts presents immense opportunities. Identifying old manuscripts, revealing lost materials quoted in later works (such as parts of the Tan\d{h}uma literature and  \textit{Mekhilta Deuteronomy}), and determining the authorship or dating of a text are all potential applications. Driven by this potential, we turn to state-of-the-art natural language processing (NLP) techniques to explore how we can leverage these tools for midrashic study.

We propose a system for the classification of rabbinic literature. This system detects unique stylistic patterns in the language of the text and can help uncover lost midrashic material quoted in later works. As a test case, we use our method to detect lost sections of the \textit{Midrash Tan\d{h}uma} that are quoted in the \textit{Yalkut Shimoni}.\footnote{A medieval midrash anthology, attributed to Simeon of Frankfort, 13th century \textsc{ce}, by the publisher of the second printed edition (Venice, 1566).}

\section{Related Work}
In recent years, advancements in natural language processing (NLP) and machine learning (ML) have greatly expanded the toolkit available for tasks such as authorship attribution, plagiarism detection, and style classification. These tools have been successfully employed in a variety of contexts, from the analysis of contemporary texts to the examination of historical documents.

In the broad context of textual analysis, \citet{Authorshipattribution} provides a thorough review of authorship attribution,
offering a comprehensive understanding of the state of the art in computational methods for authorship attribution and style classification and their applicability to different types of texts. 

The application of these techniques to biblical texts has seen particular innovation in recent years. 
\citet{10.15699/jbl.1342.2015.2754} introduced a method for automatic biblical source criticism, examining preferences among synonyms and other stylistic attributes;
technical details may be found in \citet{koppel-etal-2011-unsupervised}.
This approach laid a foundation for using stylistic analysis in the context of classical Hebrew texts.
Building on that work, \citet{Akiva2013AGU} developed an unsupervised algorithm for decomposing multi-author documents, further reinforcing the applicability of NLP and ML models in the field of authorship attribution.

\citet{ReconstructionoftheMekhiltaDeuteronomyUsingPhilologicalandComputationalTools} applied computational tools to reconstruct \textit{Mekhilta Deuteronomy}, a lost midrash halakha from the school of Rabbi Akiva. Although their research shares a common goal with ours, their approach begins with a list of candidate texts and primarily focuses on eliminating quotes or near-quotes of existing material from other sources. In contrast, our work addresses the problem of generating an initial candidate list for a specific genre.

From a methodological perspective, it is worth noting that our research also bridges the typically separate approaches of text reuse and stylometry or style detection. While text reuse predominantly focuses on content words and semantics, stylometry often concentrates on function words and other habitual linguistic choices that an author may use subconsciously. By combining these two methods, our study offers a unique perspective on text analysis. \citet{quantitativeintertextuality} provide an insightful discussion on this topic, while surveying the various computation tools used for the study of intertextuality.

Finally, the synergy between technology and humanities research is increasingly appreciated. A significant example is the Ithaca project \citep{Assael2022}, a joint venture between DeepMind and the University of Oxford, which focuses on the restoration and classification of ancient Greek epigraphs. This project demonstrates the potential of such interdisciplinary collaboration and offers a model for similar initiatives. This collaboration model resonated with our approach and influenced how we conducted our research.

\section{Method}
\begin{figure*}[t]
  \centering
    \includegraphics[width=150mm]{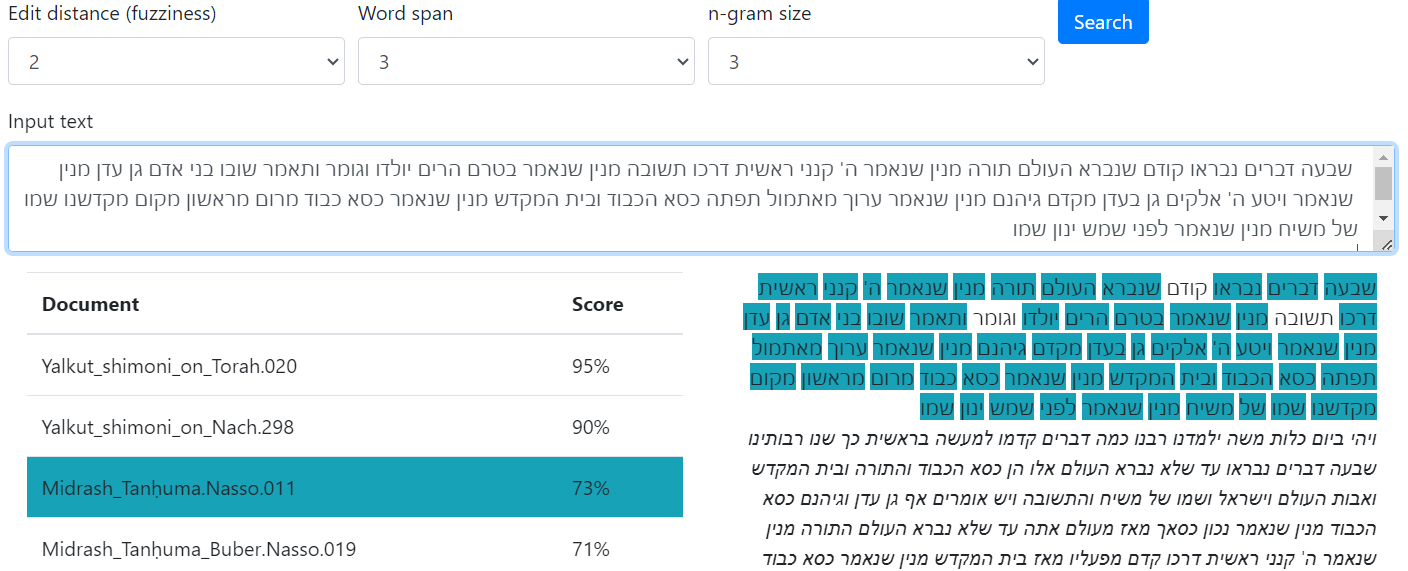}
      \caption{The text-reuse engine, RWFS, shows how a medieval midrash paragraph is reusing early material from various sources including \textit{Midrash Tan\d{h}uma}.}
    \label{fig:rwfs}
\end{figure*}

\subsection{Dataset}
Our training dataset was extracted from Sefaria's resources.\footnote{\url{https://github.com/Sefaria/Sefaria-Export}.} We use the raw text files and divide them into the following categories of rabbinic texts:

    \paragraph{Mishnah}  In this category we include all tractates of the Mishnah and the Tosefta. Both collections are generally dated to the second century \textsc{ce} and consist of rabbinic rulings and debates, organized by topic.
    
      \paragraph{Midrash Halakhah}  These  collections are dated to around the same time of the Mishnah, but they are organized according to the Pentateuch and focus more on the exegesis of biblical verses. In this class we include: \textit{Mekhilta d'Rabbi Yishmael}, \textit{Mekhilta d'Rashbi}, \textit{Sifra}, \textit{Sifre Numbers}, and \textit{Sifre Deuteronomy}.
    
     \paragraph{Jerusalem Talmud}  We include all tractates of the Jerusalem Talmud, omitting the Mishnah passages that provide the basis for discussion. These texts for the most part are written in a mixture of Hebrew and Palestinian Aramaic and are roughly dated to the 4th c.\@ \textsc{ce}.
    
      \paragraph{Babylonian Talmud}  We include all tractates of the Babylonian Talmud, omitting the Mishnah passages that provide the basis for discussion. These texts for the most part are written in a mixture of Hebrew and Babylonian Aramaic and are roughly dated to the 5th c.
    
      \paragraph{Midrash Aggadah}  In this category we include early midrash works assumed to have been composed during the amoraic period (up to the 5th c.) or slightly later. The works included in training are: \textit{Genesis Rabbah}, \textit{Leviticus Rabbah}, and \textit{Pesikta de-Rav Kahanna}. Like midrash halakhah these works follow the order of verses in the Bible, but in contrast they focus less on deriving rulings (halakhah) and more on expounding on the biblical narrative.
    Other works that we did not use during training but which we partially associate with this category include: \textit{Ruth Rabbah}, \textit{Lamentations Rabbah}, and \textit{Canticles Rabbah}.
    
      \paragraph{Midrash Tan\d{h}uma}  In this category we include later midrashic works that make up what is referred to as Tan\d{h}uma-Yelammedenu Literature.
    The works included in training are: \textit{Midrash Tan\d{h}uma}, \textit{Tan\d{h}uma Buber}, and \textit{Deuteronomy Rabbah}.
    Other works that we did not use during training but we partially associate with this category include \textit{Exodus Rabbah} starting from Section 15\footnote{See ``Exodus Rabbah,'' \textit{Encyclopaedia Judaica}, for the rationale behind this division.} and \textit{Numbers Rabbah} starting from Section 15.\footnote{See ``Numbers Rabbah,'' \textit{Encyclopaedia Judaica}, for the rationale behind this division.}
    

We divide these works into continuous blocks of 50 words. We then clean the text by removing vowel signs, punctuation and metadata. In order to neutralize the effect of orthography differences, we also expand common acronyms and standardize spelling for common words and names.

After cleaning and normalizing the data, we split our dataset into training (80\%) and validation (20\%) sets. Finally, we downsample all majority classes in the validation set to get a balanced dataset.

\subsection{Models}
\paragraph{Baseline}
For our baseline model we use a logistic regression model over a bag of $n$-grams encoding. We include unigrams, bigrams, and trigrams. We use the default parameters from scikit-learn \citep{scikit-learn} but set \texttt{\small fit\_intercept=False} to reduce the impact of varying text length and set \texttt{\small class\_weight="balanced"} to deal with class imbalance in the training data.
This type of model is highly interpretable, enabling us to see the features associated with each class. Finally, we choose this model as our baseline as it generally achieves reasonable results without the need to tune hyperparameters.

\paragraph{AlephBERT}
The next model we evaluate is AlephBERT \citep{seker-etal-2022-alephbert} -- a Transformer model trained with the masked-token prediction training objective on modern Hebrew texts. While this model obtains state-of-the-art results for various tasks on modern Hebrew, performance might not be ideal on rabbinic Hebrew, which differs significantly from modern Hebrew.
We train the pretrained model on the downstream task using the Huggingface Transformers framework \citep{wolf-etal-2020-transformers} for sequence classification, using the default parameters for three epochs.

\paragraph{BEREL}
The third model we evaluate is BEREL \citep{https://doi.org/10.48550/arxiv.2208.01875} -- a Transformer model trained with a similar architecture to that of BERT-base \citep{devlin-etal-2019-bert} on rabbinic Hebrew texts. 
In addition to the potential benefit of using a model that was pretrained on similar texts to those of the target domain, BEREL also uses a modified tokenizer that does not split up acronyms that would otherwise be interpreted as multiple tokens with punctuation marks in between. (Acronyms marked by double apostrophes [or the like] are very common in rabbinic Hebrew.) We train the pretrained model on our downstream task in an identical fashion to the training of the AlephBERT model.

\paragraph{Morphological}
Finally, we also train a model that focuses only on morphological features in the text, in an attempt to neutralize the impact of content words. We expect this type of model to detect more ``pure'' stylistic features that help discriminate between the different textual sources.
To extract features from the text, we use a morphological engine for rabbinic Hebrew created by DICTA.%
\footnote{\url{https://morph-analysis.dicta.org.il/}.} 
We then train a logistic regression model over an aggregation of all morphological features that appear in a given paragraph.

\subsection{Text Reuse Detection}

To achieve our end goal of detecting lost Tan\d{h}uma material, we combine our style classification model with a filtering algorithm based on text reuse detection. 

For reuse detection, we utilize RWFS (Rolling Window Fuzzy Search) by \cite{schor2021digital}.%
\footnote{This intertext engine was designed for this purpose by our partners at eLijah Lab, University of Haifa (\url{https://elijahlab.haifa.ac.il/about-eng}).}
RWFS uses fuzzy full-text search on windows of $n$-grams. The system is built on top of a Lucene index,\footnote{\url{https://lucene.apache.org/core}.} and uses a web-based interface to provide easy querying to technological and non-technological users. 

For our corpus of texts for this engine we use all biblical and early rabbinic works using the texts available on Sefaria. We use 3-gram matching and permit a Levenshtein distance of up to 2 for each individual word in the $n$-gram. The match score for each retrieved document is given by the number of $n$-gram matches divided by the length of the query and the results are sorted accordingly (Figure \ref{fig:rwfs}).

\subsection{Detecting Lost Tan\d{h}uma Candidates}
\label{subsection:lost}
Tan\d{h}uma-Yelammedenu Literature is a name given to a genre of late midrash works, some of which are lost and only scarcely preserved in anthologies or Genizah fragments (\citealp{bregman_2003, nikolsky_atzmon_2021}). One of the lost works was called ``Yelammedenu,'' and we know about it since it is cited in various medieval rabbinic works such as \textit{Yalkut Shimoni} and the \textit{Arukh}.\footnote{An 11th century  dictionary of rabbinic literature by Nathan ben Jehiel of Rome.} While lost Tan\d{h}uma material is explicitly cited in some works, it is often quoted without citation in other midrash anthologies.

To find candidates for ``lost'' Tan\d{h}uma passages, we apply the following process:
\begin{enumerate}\itemsep0em 
    \item Extract all passages from the given midrash collection, in our case  \textit{Yalkut Shimoni}.
    \item Split long passages into segments of up to 50 words.
    \item Run these segments through the style detection model.
    \item Collect segments for which our model gives the highest score to the Tan\d{h}uma class.
    \item Run these segments through a text-reuse engine.
    \item Keep only segments that do not have a well established source. (Our threshold was $\#n$-gram matches $\le$ $0.2\times\#n$-grams in query.\footnote{Subsequent experiments using different study cases demonstrated the need to adjust the threshold based on the specific task.})
    
\end{enumerate}

\section{Results}

\begin{figure*}[t]
    \includegraphics[trim=0 0 0 100,clip,width=\linewidth]{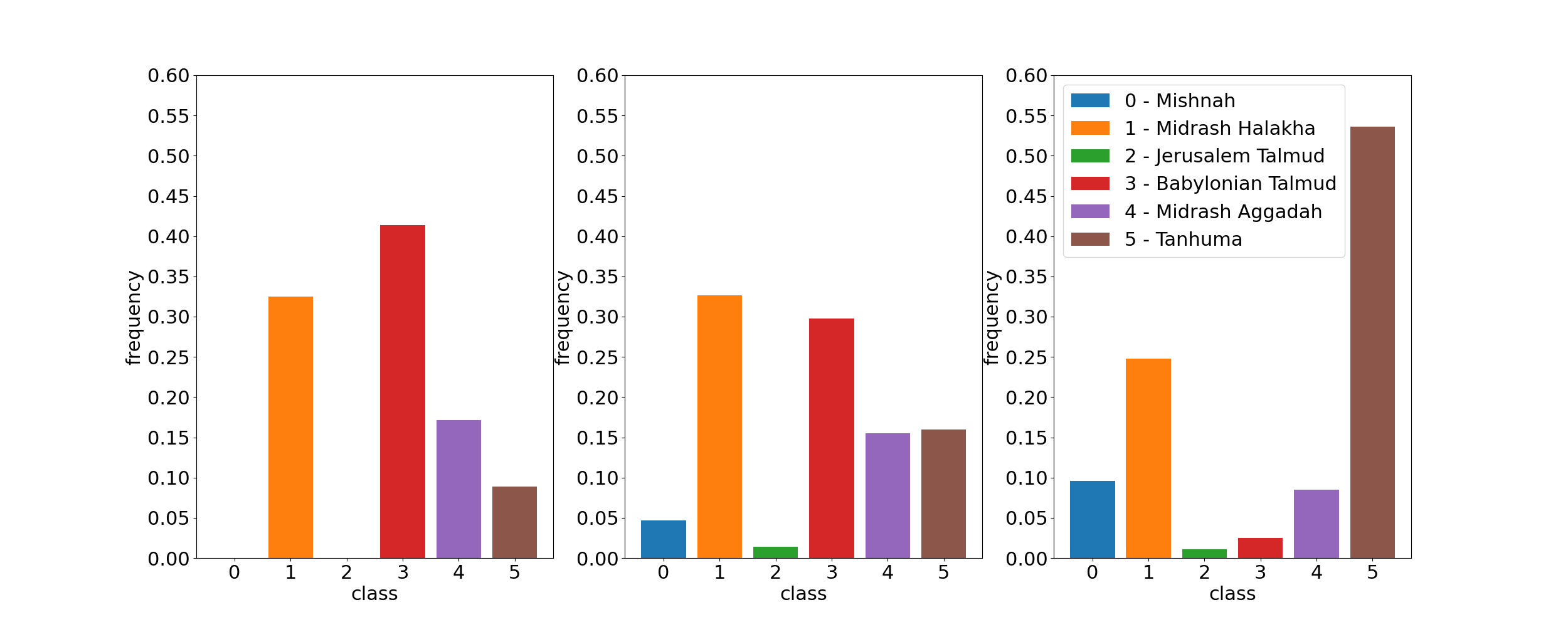}
      \caption{From left to right: (1) class frequencies for  passages based on text reuse detection in \textit{Yalkut Shimoni}; (2) predicted class frequencies for passages with high text reuse score; (3) predicted  frequencies for passages with low  reuse score. }
    \label{fig:freq}
\end{figure*}

\begin{table}
  \newcolumntype{+}{>{\global\let\currentrowstyle\relax}}
  \newcolumntype{^}{>{\currentrowstyle}}
  \newcommand{\rowstyle}[1]{\gdef\currentrowstyle{#1}%
    #1\ignorespaces
  }

  \centering
  \begin{tabular}{+>{\bfseries}l^c^c}
    \hline
    \rowstyle{\bfseries}
    & Validation Acc \\
    Baseline & 0.867 \\
    AlephBERT & 0.879 \\
    BEREL & \textbf{0.922} \\
    Morphological & 0.560 \\
    \hline
  \end{tabular}

  \caption{Model accuracy on validation set.}
  \label{table:accuracy}
\end{table}

As can be seen in Table \ref{table:accuracy}, our baseline model achieves well over the random guess accuracy of 0.166 on the validation set, and achieves almost the same accuracy as the Aleph\-BERT fine-tuned model. The BEREL-based model leads by a significant margin. However, we encountered multiple challenges when using this model for inference on paragraphs from \textit{Yalkut Shimoni}:
\begin{enumerate}
    \item The model's scores were not calibrated, most predictions were very close to 1.0 or 0.0, making it hard to experiment with different thresholds.
    \item BEREL's accuracy on a small sample of paragraphs from \textit{Yalkut Shimoni} was significantly lower than the corresponding validation accuracy. It seems that BEREL might have relied on some orthographic features that appeared in the training and validation sets but not in the new out-of-distribution text.
    \item Transformer-based models are generally less interpretable, and have higher inference costs than classical ML models such as logistic regression.
\end{enumerate}

For these reasons, we decided to use our baseline model for inference on \textit{Yalkut Shimoni}.

\begin{figure}[t]
    \centering
    \includegraphics[width=120mm]{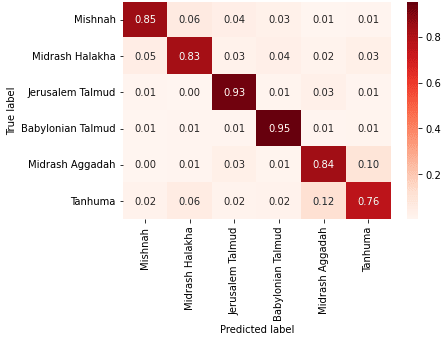}
      \caption{Confusion matrix for baseline model, normalized by row.}
    \label{fig:confusion}
\end{figure}

In Figure \ref{fig:confusion}, we can see that the the most common errors are mixing `Tan\d{h}uma' with `Midrash Aggadah' which are indeed relatively similar genres. On the other hand, `Babylonian Talmud' and `Jerusalem Talmud' seem to be the most distinct classes, perhaps due to their extensive use of Aramaic in addition to Hebrew, each in its own unique dialect.

After taking the whole \textit{Yalkut Shimoni} on the Pentateuch and following the process described in Section \ref{subsection:lost}, we can analyze the prevalence of each class in the collection. As can be seen in Figure \ref{fig:freq}, the Babylonian Talmud is the most quoted class, while the Jerusalem Talmud is rarely, if ever, quoted. Our classifier gives a similar distribution to that of the text-reuse engine. However, when looking only at passages with low reuse score we see that the Babylonian Talmud rarely appears while `Tan\d{h}uma' becomes the most frequent predicted class by far, followed by `Midrash Halakha.' This aligns with the fact that we know of lost works that belong to these categories, while the Babylonian Talmud was well preserved throughout the generations as the core text of the rabbinic tradition. 

To evaluate our classifier on the target task, we sampled a random set of 50 items classified as Tan\d{h}uma for manual labeling. 
A midrash expert analyzed these passages and looked them up in the early print edition of \textit{Yalkut Shimoni}, which tends to include citations in the margins. Sections that were ascribed to Yelammedenu 
and sections that were recognized as being typical Tan\d{h}uma material were labeled as ``positive,'' while all other passages were labeled ``negative.'' 
Out of these items, 22 were cited as Yelammedenu, while an additional 8 were recognized as typical Tan\d{h}uma material from lost sources,\footnote{These latter items are perhaps the more exciting find as they have previously been unidentified.} yielding an approximate precision of 60\%. 

\begin{figure}[t]
    \centering
    \includegraphics[width=120mm]{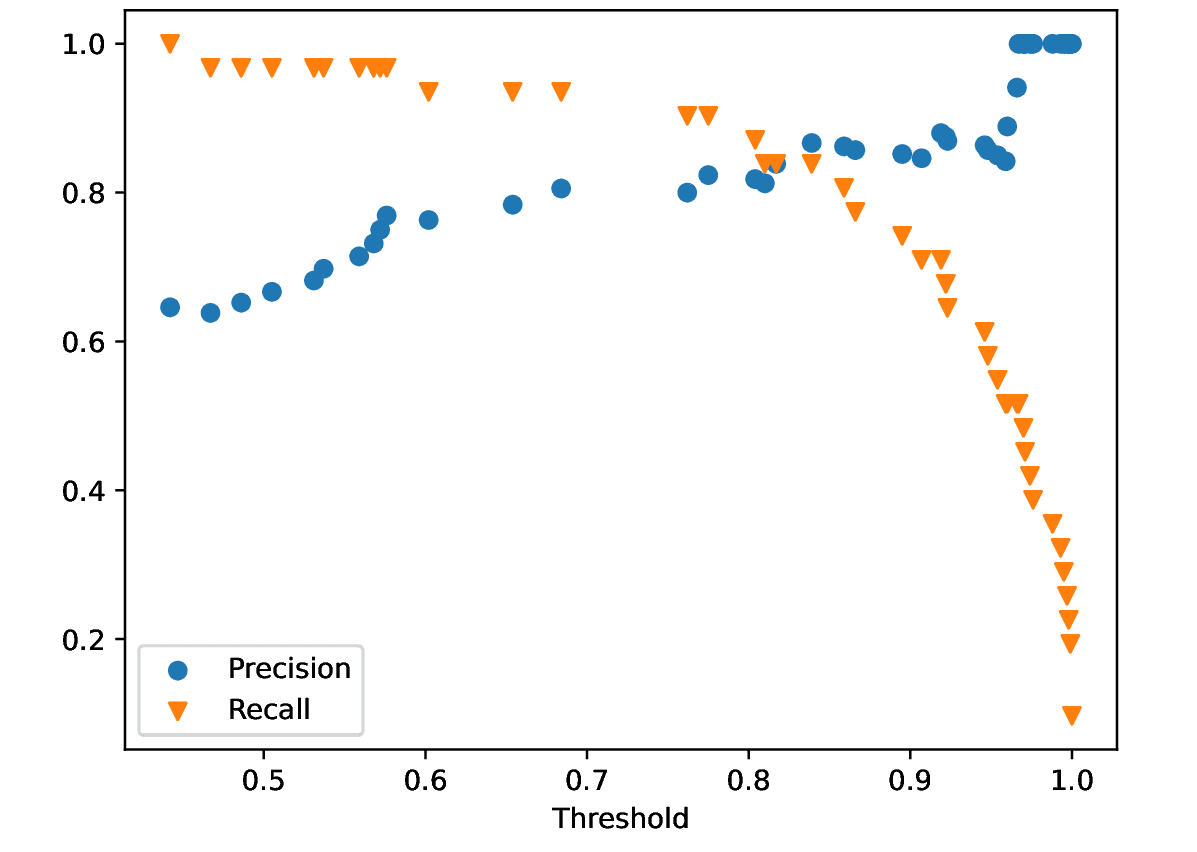}
      \caption{Precision and recall as function of the decision threshold for lost Tan\d{h}uma material.}
    \label{fig:precision}
\end{figure}

From Figure \ref{fig:precision}, we see that the precision grows monotonically with the decision threshold, indicating that the model is useful in recovering lost Tan\d{h}uma material. Furthermore, we see that we can achieve a precision of approximately 80\% by setting an appropriate decision threshold without a high cost to  recall.

\begin{figure*}[t]
    \centering
    \includegraphics[width=150mm]{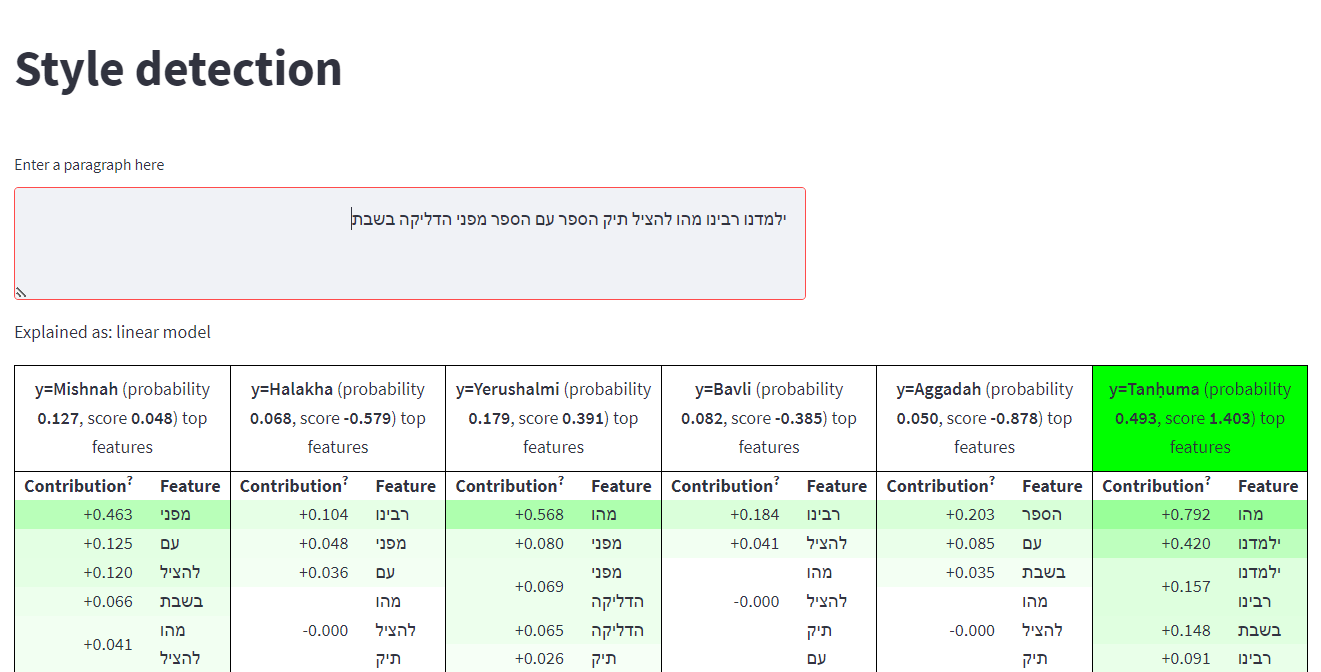}
      \caption{An example of our application's output on a typical \textit{Midrash Tan\d{h}uma} text.}
    \label{fig:style1}
\end{figure*}

\begin{figure*}[t]
    \centering
    \includegraphics[width=100mm]{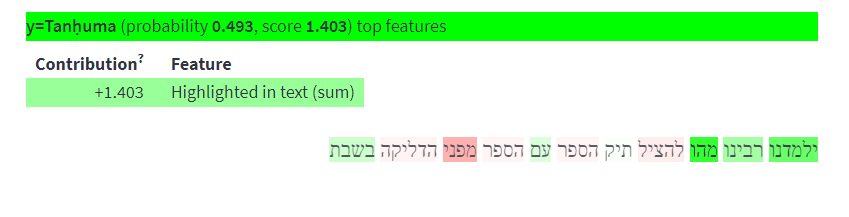}
      \caption{Significant features are highlighted in the text to provide an explanation that is easier to process.}
    \label{fig:style2}
\end{figure*}

\subsection{Findings}
Using the methodology we described to investigate thoroughly the makeup of \textit{Yalkut Shimoni} on Deuteronomy, there were some interesting findings  that arose,  as well as some questions.

A systematic expert examination of all results for the first half of Deuteronomy (approximately 10\% of the Pentateuch) revealed that all known citations of Yelammedenu, ranging from 100 to 600 words, had at least one sub-paragraph of 50 words recognized as part of the genre. In most cases, the majority of the citation was also identified as part of the genre. In practical terms, this means that every block of text was correctly identified, resulting in a 100\% recall rate.

Interestingly, one paragraph that was detected as ``lost Tan\d{h}uma'' material was actually cited as \textit{Deuteronomy Rabbah} in the early print version of \textit{Yalkut Shimoni}. However, 
 our version of \textit{Deuteronomy Rabbah} had a very low text reuse match for this paragraph. This result raises the question of whether the author of \textit{Yalkut Shimoni} had a different version of the text from what we have.%
 \footnote{We do know of one alternative version to the text that was prevalent in Spain in the 13th c. This is known as ``Deuteronomy Rabbah [Lieberman].''}

 Another question that rises from this phenomenon is the extent to which the midrash collators rephrase and reorganize the early material they work with as opposed to copying full sections.%
 \footnote{It seems, for example, that \textit{Yalkut Shimoni} on the Prophets and the Yemenite anthology, \textit{Midrash Hagadol}, tend to rework early material more extensively than does \textit{Yalkut Shimoni} on the Pentateuch.}

Another notable finding is that some of the lost midrash collections known only from Ashkenaz (e.g. \<'bkyr>, \<'sph>, \<dbrym zw.t'>) got a very high score for Tan\d{h}uma style. This might hint that there is a stronger connection between these works and the Tan\d{h}uma literature than previously thought, and perhaps they should be considered as part of the same genre as Tan\d{h}uma in some contexts. \footnote{The strong correlation of these texts with the Tan\d{h}uma genre has been validated by the only comprehensive study of these texts, as documented by \citet{Geula}. The fact that our method has highlighted a largely unrecognized phenomenon within the Humanities field underscores its significant practical value for scholars.}

Finally, there were a number of paragraphs from \textit{Sifre Deuteronomy}, a midrash halakha of the tannaitic period, that were detected by our classifier as Tan\d{h}uma. One such paragraph (\textit{Sifre Deuteronomy} 26) contained some notable phrases associate with Tan\d{h}uma and other later midrashic works including \<zhw +s'mr hktwb> (``As it is said in Scripture'')  and \<hqdw+s brwk hw'> (``The holy one, blessed be He'').\footnote{As opposed to the prevalent use of \<hmqwm> (lit.\@ ``The Place'') in the tannaitic period, for example, as a metonym for God.} As it turns out, in one of the manuscripts (Vatican manuscript 32) some of these terms do not appear. This phenomenon might suggest that over the course of time some terms from later periods such as the Tan\d{h}uma literature might have made their way into our current versions of earlier texts.

\section{User Tools}
In order to provide access to our model's predictions and corresponding explanations, and turn our research into a tool that can assist midrash scholars, we built an interactive application based on the open-source Streamlit platform to wrap our model's inference process.
Given an input paragraph, the app displays the scores for each of the classes along with features' (unigrams, bigrams and trigrams) corresponding contributions (Figure \ref{fig:style1}).

Additionally, as can be seen in Figure \ref{fig:style2}, we display the contribution of the various parts of the text to the prediction in a more convenient way by highlighting the important features in the text.

\subsection*{Conclusion}
In conclusion, our method for detecting Tan\d{h}uma sections in \textit{Yalkut Shimoni} showcases its potential as a valuable tool for scholars involved in the recovery of lost rabbinic material. This is particularly relevant in the ongoing initiative to develop a digital library of Tan\d{h}uma-Yelammedenu Literature, where our work has significant implications. The tools and classifiers we have developed in this study will prove useful for midrash researchers engaged in compiling various Tan\d{h}uma sources and discovering related and potentially lost material of this genre. These resources are publicly accessible,\footnote{Online at \url{https://github.com/shlomota/Midrash-Style-Classification}} fostering their widespread future use in related projects.

Beyond its immediate application, our method offers potential for wider use in Jewish studies. One exciting future direction is to examine the baraitot\footnote{A tannaitic tradition not incorporated in the Mishnah, see: ``Baraita,'' \textit{The Jewish Encyclopedia}.} that appear in  the Babylonian Talmud and in the Jerusalem Talmud. Investigating their interrelationships and connections with other tannaitic sources could provide fresh insights into these traditions.

Additionally, the suggested method could be applied to the many unorganized and unstudied manuscripts discovered in collections such as the Cairo Geniza,\footnote{Online at \url{https://fgp.genizah.org}.} paving the way for their automatic classification. Despite the challenges posed by the noisy text created by current engines for handwritten text recognition, the potential benefits to the academic community in terms of improved access and understanding of these vital documents are considerable.

This research represents a significant advancement in the use of computational tools for analyzing Jewish literature and traditions. As we continue to refine and expand our methodologies, we anticipate further contributions to the discovery and development of innovative tools, which will undoubtedly enhance our understanding of Jewish textual traditions.

\bibliographystyle{plainnat}
\bibliography{jdmdh-example}

\appendix\footnotesize

\end{document}